%% file: main.tex
\documentclass[10pt,twocolumn,letterpaper]{article}

\usepackage{iccv}              % To produce the CAMERA-READY version

\definecolor{iccvblue}{rgb}{0.21,0.49,0.74}
\usepackage[pagebackref,breaklinks,colorlinks,allcolors=iccvblue]{hyperref}

\def\paperID{5226} % *** Enter the Paper ID here
\def\confName{ICCV}
\def\confYear{2025}

\title{Fuse Before Transfer: 
Knowledge Fusion for Heterogeneous Distillation}

\author{First Author\\
Institution1\\
Institution1 address\\
{\tt\small firstauthor@i1.org}
\and
Second Author\\
Institution2\\
First line of institution2 address\\
{\tt\small secondauthor@i2.org}
}

\usepackage{hyperref}
\usepackage{url}
\newcommand\blfootnote[1]{%
  \begingroup
  \renewcommand\thefootnote{}\footnote{#1}%
  \addtocounter{footnote}{-1}%
  \endgroup
}
\usepackage{multirow}
\usepackage{colortbl}
\PassOptionsToPackage{prologue,dvipsnames}{xcolor}

\makeatletter
\newcommand\figcaption{\def\@captype{figure}\caption}
\newcommand\tabcaption{\def\@captype{table}\caption}
\makeatother

\usepackage{graphicx}
\usepackage{booktabs}

\usepackage{xcolor}
\usepackage[normalem]{ulem}
\usepackage{soul}
\usepackage{amsmath}

\soulregister\cite7
\soulregister\ref7 
\soulregister\cref7 
\soulregister\citep7 % 针对\citep命令
\soulregister\citet7 % 针对\citet命令
\soulregister\pageref7 % 针对\pageref命令

\hypersetup{colorlinks=true,breaklinks,linkcolor=red,citecolor=goodblue}
\usepackage[pagebackref,breaklinks,colorlinks,allcolors=iccvblue]{hyperref}
\definecolor{goodblue}{HTML}{0071bc}

\AtEndPreamble{
    \usepackage[capitalize]{cleveref}
    \crefname{section}{Sec.}{Secs.}
    \Crefname{section}{Section}{Sections}
    \crefname{table}{Tab.}{Tabs.}
    \Crefname{table}{Table}{Tables}
}

\newcommand{\authorskip}{\hspace{4.8mm}}
\author{Guopeng Li\textsuperscript{1*}\hspace{-1.em}
\authorskip Qiang Wang\textsuperscript{3} \hspace{-1.em}
\authorskip Ke Yan\textsuperscript{3} \hspace{-1.em}
\authorskip Shouhong Ding\textsuperscript{3}
\hspace{-1.em}
\authorskip Yuan Gao\textsuperscript{2}$^\dag$ \hspace{-1.em} 
\authorskip Gui-Song Xia\textsuperscript{2}$^\dag$ \\[2mm]
\textsuperscript{1}School of Computer Science, \textsuperscript{2}School of Artificial Intelligence, Wuhan University \\
\textsuperscript{3}Tencent YouTu Lab \\
\texttt{\{guopengli, guisong.xia\}@whu.edu.cn}, \texttt{ethan.y.gao@gmail.com} \\
\texttt{\{albertqwang, kerwinyan, ericshding\}@tencent.com} 
}

\begin{document}

\maketitle
\blfootnote{$^*$ Work done during internship at Tencent YouTu Lab.
\\
\indent \indent $^\dag$ indicates the corresponding authors.}

\vspace{-0.7cm}
\begin{abstract}
Most knowledge distillation (KD) methods focus on teacher-student pairs with \textbf{similar architectures}, such as both being CNN models. The potential and flexibility of KD can be greatly improved by expanding it to Cross-Architecture KD (CAKD), where the knowledge of homogeneous and \textbf{heterogeneous} teachers can be distilled selectively. However, substantial feature gaps between heterogeneous models (\eg, ViT teacher \emph{v.s.} CNN student) make CAKD extremely challenging, caused by the distinction of inherent \textcolor{blue}{inductive biases} and \textcolor{red}{module functions}. To this end, we fuse heterogeneous knowledge before transferring it from teacher to student. This fusion combines the advantages of both cross-architecture inductive biases and module functions by \textcolor{blue}{merging different combinations of convolution, attention, and MLP modules} \textcolor{red}{derived directly from student and teacher module functions}. Furthermore, heterogeneous features exhibit diverse spatial distributions, hindering the effectiveness of conventional pixel-wise MSE loss. Therefore, we replace it with a spatial-agnostic InfoNCE loss. Our method is evaluated across various homogeneous models and arbitrary heterogeneous combinations of CNNs, ViTs, and MLPs, yielding promising performance for distilled models with a maximum gain of 11.47\% on CIFAR-100 and 3.67\% on ImageNet-1K. Code is available at \url{https://github.com/liguopeng0923/FBT}.

\end{abstract}

\begin{figure}[!t]
  \centering
  \includegraphics[width=.7\linewidth]{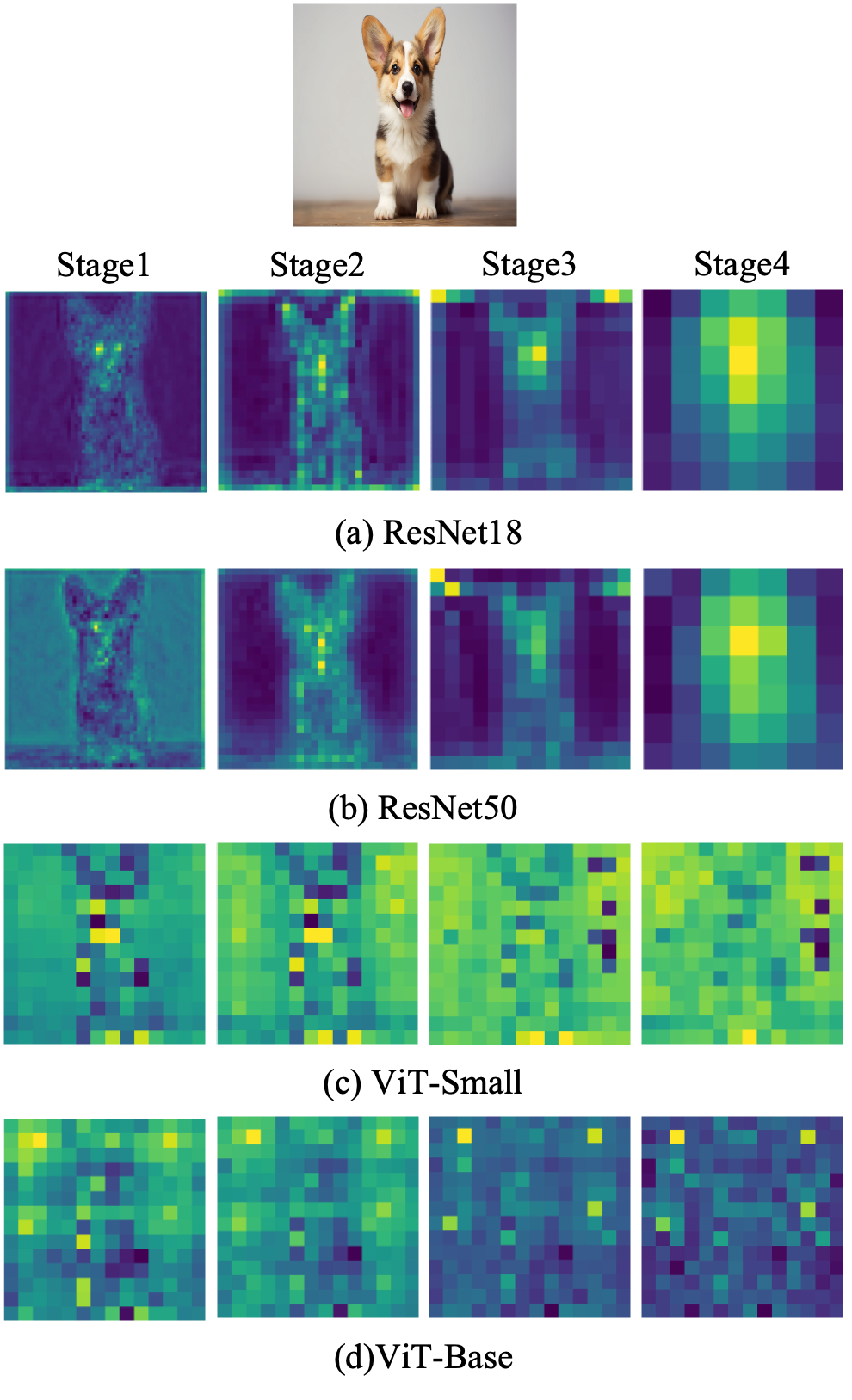}
\vspace{-0.2cm}
  \caption{\textbf{Significant feature gaps among heterogeneous models.} For example, (a) CNN-based models~\cite{ResNet} and (c) ViT-based models~\cite{vit} have different features in different stages caused by different inductive biases and module functions.
  }

  \label{fig:teaser}
\end{figure}
\section{Introduction}
\label{sec:intro}

\begin{figure*}[!t]
\centering
\begin{minipage}{\textwidth}
\centering
\includegraphics[width=\linewidth]{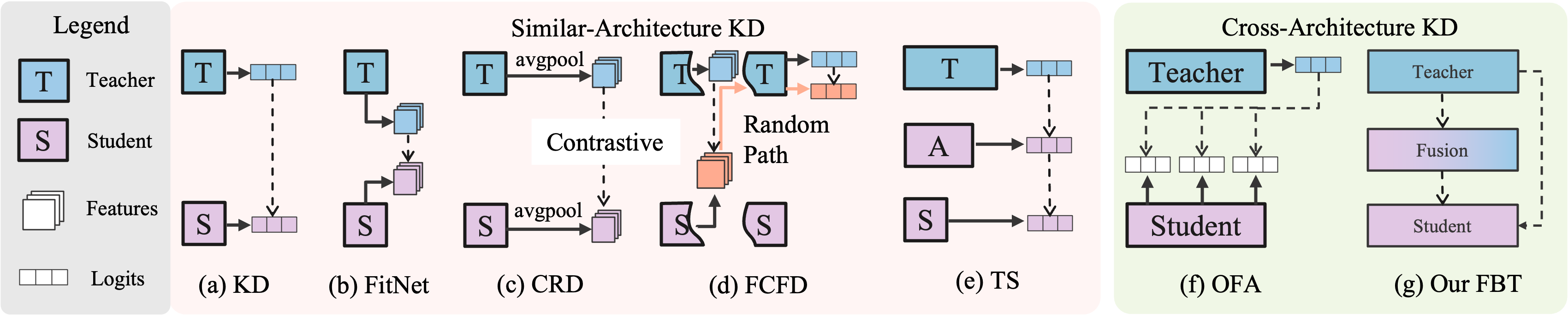}
\end{minipage}
\begin{minipage}{\textwidth}
\centering
\resizebox{\textwidth}{!}{
\centering

\begin{tabular}{c|c|c|c|c|c|c|c}
\hline \hline
                          & KD~\cite{KD}          & FitNet~\cite{FitNet}           & CRD~\cite{CRD}              & FCFD~\cite{Function}          & TS~\cite{progressive}          &OFA~\cite{OFA}             & Ours \\ \hline
                                                 
Knowledge to Distill                      & Logits    & Feature   & Feature   & Feature & Logits  & Logits   & \textbf{Feature}             \\
Generic                  & Yes              & No              & Yes             & No       & No       & Yes             & \textbf{Yes}                       \\
Scheme                     & T.-S. & T.-S. & T.-S. & T.-S. & T. $\rightarrow$ A. $\rightarrow$ S. & T.-S. & \textbf{T.-F.-S.} \\
Heterogeneous Inductive Bias Fusing  & No              & No              & No              & No         & No     & No              & \textbf{Yes}                       \\
Heterogeneous Module Merging & No              & No              & No              & Yes  & No   & No              & \textbf{Yes}                       \\
Training Cost            & \textbf{Very Low}        & Middle          & High            & High      & Middle      & Middle          & Low                       \\
Loss Function             & $\mathcal{L}_{\mathrm{KL}}$              & $\mathcal{L}_{\mathrm{KL}}$ + $\mathcal{L}_{\mathrm{MSE}}$          & $\mathcal{L}_{\mathrm{InfoNCE}}$         & $\mathcal{L}_{\mathrm{KL}}$ + $\mathcal{L}_{\mathrm{MSE}}$   & $\mathcal{L}_{\mathrm{KL}}$      & $\mathcal{L}_{\mathrm{OFA}}$             & $\mathcal{L}_{\mathrm{OFA}}$ + $\mathcal{L}_{\mathrm{InfoNCE}}$    \\ \hline \hline          
\end{tabular}}
\end{minipage}
\vspace{-0.2cm}
\figcaption{\label{tab:Comparisions} \textbf{The taxonomy of our method.} Our methods are feature-based, generic, and three-level, which fuses heterogeneous inductive biases and module functions with an efficient fused model. Target-wise $\mathcal{L}_{\mathrm{OFA}}$ and spatial-agnostic $\mathcal{L}_{\mathrm{InfoNCE}}$ are more suitable for CAKD than $\mathcal{L}_{\mathrm{KL}}$ and $\mathcal{L}_{\mathrm{MSE}}$. To the best of our knowledge, our FBT is one of the pioneer works in feature-based generic distillation.}\vspace{-0.3cm}
\end{figure*}

Knowledge Distillation (KD)~\citep{KD,FitNet} has been demonstrated as a powerful method to transfer knowledge from a cumbersome teacher to a compact student. Compared to the model trained from scratch, the performance of distilled students usually improves significantly. Generally, knowledge transferred is derived from output logits (logits-based KD~\citep{sun2024logit}) or intermediate features (feature-based KD~\citep{FitNet}) of the teacher model. Therefore, it is intuitive to understand different teachers have different knowledge (logits or features) determined by their unique architectures~\citep{multi-teacher}.

Most existing KD methods focus on similar-architecture distillation~\citep{FitNet,CRD,Function} (called SAKD), \ie, optional teachers are restricted to a limited scope with structures similar to the student model. This presents two principal limitations: \textbf{(1) Limited Potential:} Compared to the broader range of arbitrary teachers (including homogeneous and heterogeneous ones), the restricted scope of teachers in SAKD may fail to include the optimal knowledge. For instance, as OFA~\citep{OFA} demonstrated, distilling knowledge from a heterogeneous ViT-Base to ResNet50 yields better performance compared to using a ResNet152 as the homogeneous teacher. \textbf{(2) Limited Flexibility:} The emergence of new models~\citep{liu2022convnet,mixer} or the scarcity of homogeneous teachers in domain-specific tasks~\citep{u-net,li2024unleashing} poses significant challenges in obtaining suitable homogeneous teachers, thereby impeding the applicability of SAKD. Thus, this paper tends to expand KD to cross-architecture KD (CAKD), broadening the scope of optional teachers and thus improving the potential and flexibility of KD~\citep{KD,multi-teacher,cao2023learning}.

% Although some methods make efforts to distill the knowledge between CNNs and ViTs~\citep{zhao2023cumulative}, they are a small part of our CAKD and are complex to be applied to other teacher-student pairs. Besides, while some logits-based methods can be easily applied to CAKD ~\citep{KD,sun2024logit,OFA}, they are not optimal because they ignore the importance of feature-based knowledge.

In CAKD, the main challenge is that heterogeneous teachers and students have significant representation gaps, detailed in CKA analyses of OFA~\citep{OFA} and feature maps in \cref{fig:teaser}. These gaps stem from inherent differences in inductive biases~\citep{dovitlikecnn} and module functions~\citep{Function}. \textbf{(1) Inductive biases:} As demonstrated in ~\cite{dovitlikecnn,howvitworks}, convolutional neural network-based models (CNNs)~\citep{ResNet,MobileNet} exhibit locality and translation-equivariance, while multi-head-self-attention-based models (MSAs)~\citep{vit} and multilayer-perception-based models (MLPs)~\citep{mixer} depend on patchify and long-distance dependency. Consequently, CNN-generated features are located in local objects in \cref{fig:teaser} (a,b), but most MSA/MLP models generate global features in \cref{fig:teaser} (c,d). \textbf{(2) Module functions:} Varied module functions generate different features at different stages. For instance, features of shallow and deep layers in ViT have higher similarity than hierarchical CNN~\citep{howvitworks,dovitlikecnn} in \cref{fig:teaser}(a,c). 

% While OFA~\citep{OFA} attempts to address heterogeneous feature gaps by projecting features into the logits space, it is suboptimal due to substantial damage to feature-specific knowledge. 
% This dilemma leads to a natural question: \emph{Can we get the best use of different inductive biases and module functions, thereby reducing heterogeneous feature gaps in CAKD?}
% Therefore, heterogeneous T.-S. pairs in CAKD exhibit significantly different inductive biases and module functions compared to SAKD, leading to substantial representation gaps that impede effective feature transfer. 

To alleviate feature gaps in CAKD, as shown in ~\cref{tab:Comparisions} (g), we \textbf{F}use heterogeneous modules \textbf{B}efore \textbf{T}ransferring (FBT) by merging sequentially CNN/MSA/MLP modules derived from teachers and students. The archiercture of fused model is adaptive according to distilled model pairs.

Our FBT is well-motivated by the following popular beliefs: \textbf{(1) How do we fuse heterogeneous inductive biases?} As demonstrated in ~\citep{howvitworks,li2023convmlp,li2023uniformer}, CNNs and MSAs/MLPs are complementary. A fused model that uses CNNs in the early stages and MSAs/MLPs in the later stages can benefit from both local and global inductive biases. Compared to existing KD like \cref{tab:Comparisions} (a-f), our fused model in \cref{tab:Comparisions} (g) merges 
 CNN and MSA/MLP modules, thereby reducing distillation gaps attributed to inductive biases. 
 \textbf{(2) How do we fuse heterogeneous module functions?} As demonstrated in ~\citep{Function,reviewkd}, the disparity between heterogeneous features is also from different module functions, \ie, how the models will read, decode, and process the inputs. Therefore, a fused model comprising student and teacher modules not only optimizes the functional similarity~\citep{Function} between heterogeneous T.-S. pairs\footnote{\noindent This paper shorten the teacher, fusion, and student by T., F., and S.}, but also introduces minimal additional learnable parameters. 
\textbf{(3) How do we align heterogeneous features spatially?} Widely used MSE loss aligns the pixel-wise features, which is inadequate for spatially diverse heterogeneous features. For example, (a) and (c) in \cref{fig:teaser} show distinct spatial distributions at four stages, which are hard to align pixel by pixel. To address this, we apply average pooling to smooth the spatial of features and utilize a spatial-agnostic loss~\citep{he2020momentum} to align heterogeneous features. 

In view of the above analysis, the taxonomy of our FBT is illustrated in ~\cref{tab:Comparisions}. Our FBT falls under the category of \emph{feature-based methods} for \emph{generic distillation} with an \emph{adaptive fusion} scheme. In our experiments, the proposed FBT greatly enhances the performance of student models in both CAKD and SAKD, achieving a maximum gain of 11.47\% on the CIFAR100 and 3.67\% on the ImageNet-1K.

\section{Releated Work}
\subsection{Taxonomy of our methods}

As shown in \cref{tab:Comparisions}, the majority of existing KD methodologies concentrate on homogeneous distillation by using a single projector (\eg, single linear layer) to align the output logits~\citep{KD,DIST,sun2024logit}, intermediate features~\citep{FitNet,reviewkd,Function}, feature embeddings~\citep{CRD}, and module functions~\citep{Function} of T.-S. pairs, thanks to the highly-similar features between homogeneous T.-S. pairs. However, they fall short in addressing the complexities of heterogeneous distillation, where the distinct features between heterogeneous T.-S. pairs pose significant challenges. Although OFA~\cite{OFA} and ~\cite{liu2022cross} achieve consistent improvements for arbitrary T.-S. pairs, it does so at the expense of sacrificing feature information. 

Additionally, several other works are pertinent to our method: (1) We note that some methods attempt to distill the knowledge between CNNs and MSAs~\citep{zhao2023cumulative}, but they are tailored to specific T.-S. pairs rendering them impractical for our arbitrary T.-S. CAKD. (2) Certain methods apply progressive distillation to transfer the knowledge via a middle model~\citep{progressive,cao2023learning,multi-teacher}, but they are progressive training strategies that are not training algorithms designed for transferring knowledge between heterogeneous T.-S. pairs. (3) While some logits-based methods can be easily applied to CAKD ~\citep{KD,sun2024logit,OFA}, they are suboptimal because they overlook the significance of feature-based knowledge~\citep{FitNet}. (4) Some works~\cite{liu2022cross,wu2024aligning} attempt to align the heterogeneous features, but they also ignore the basic heterogeneous gaps in different inductive bias and module functions.

In this paper, our method dives into the nature of heterogeneous feature gaps (\ie, caused by inductive bias and module functions) and introduces a simple fusion strategy to facilitate smoother feature transfer between heterogeneous T.-S. pairs. We hope this adaptive knowledge fusion strategy motivates more work in heterogeneous fusion.

\subsection{Hybrid Model}
% \noindent \textbf{Knowledge Distillation.}

% Most KD methods work under the assumption of similar architectures for both student and teacher models, restricting their power to address the substantial feature discrepancies between cross-architecture models. Therefore, how to transfer the feature knowledge between cross-architecture models is still a meaningful yet challenging problem.

As illustrated in \cref{fig:teaser}, different models exhibit different features caused by different inductive biases and module functions. ~\citep{dovitlikecnn} investigates the internal representation structures of ViT and CNN models, revealing significant differences between their heterogeneous features. ~\citep{howvitworks} further provides some fundamental explanations for this phenomenon. Specifically, CNNs are data-agnostic and channel-specific high-pass filters, while MSAs are data-specific and channel-agnostic low-pass filters. Therefore, 
researchers~\citep{howvitworks} think CNNs and MSAs are complementary, which inspires them to design a hybrid model following the rules of ``alternately replacing CNN blocks with MSA blocks from the end of a baseline CNN model''. The hybrid model outperforms CNNs in both large and small data regimes~\citep{howvitworks}. Furthermore, the architecture of MLP models~\citep{mixer} is notably similar to ViTs not CNNs, so ConvMLP~\citep{li2023convmlp} also achieves advanced performance in basic visual tasks by the co-design of CNNs and MLPs. In a nutshell, hybrid CNN-MSA/MLP models improve performance and efficiency through the combination of different inductive biases and module functions.

Inspired by the design of the hybrid model, we mitigate the heterogeneous feature gaps by fusing heterogeneous knowledge between cross-architecture T.-S. pairs.

\section{Method}
\begin{figure*}[!t]
  \centering
\includegraphics[width=.9\linewidth]{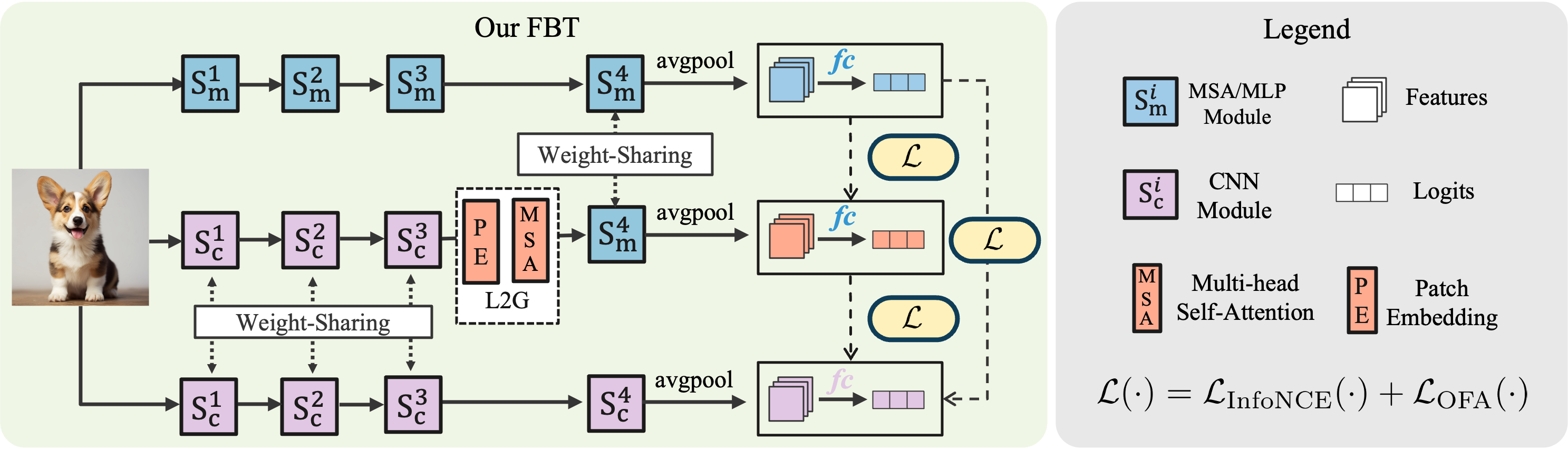}
  \caption{\textbf{Overall.} Firstly, our FBT fuses heterogeneous knowledge by merging the first three stages of CNNs, a projector L2G, and the last stage of MSAs/MLPs into the fused model. The fused model is adaptive and can be adjusted automatically according to different T.-S. pairs. Secondly, supervised by spatial-agnostic INFONCE loss~\cite{he2020momentum} and target-wise OFA loss~\cite{OFA}, the knowledge is transferred from the teacher to the fused model and student. All models are split into four stages following ~\citep{OFA}. \label{fig:overall}}
\end{figure*}

\subsection{Preliminaries\label{Preliminaries}}
Existing KD methods perform well in homogeneous distillation, but they may fail in heterogeneous teachers and students. The primary reasons stem from fundamentally distinct feature and logit spaces, caused by different \textit{inductive biases} and \textit{module functions} of heterogeneous models.

\noindent \textbf{Inductive Bias and module functions.\label{changellengings}} Inductive bias refers to the set of assumptions that a model uses to make predictions on unseen data~\citep{ren2022co}. Module functions describe how a model reads, encodes, decodes, and processes the data~\citep{Function}. Heterogeneous models exhibit different inductive biases and module functions. (1) CNN models~\citep{ResNet} slide a set of learnable local kernels across the pixel-level image, focusing on local receptive fields.  The weight-sharing kernels are applied across the entire image, providing the network with translation-equivariance to recognize an object regardless of location. (2) MSA models~\citep{vit} split the input image into patches, and attention modules calculate the scores between the Query and Key to generate attention maps. This process, capturing long-distance dependency, allows the model to consider the global information from all patches. (3) MLP models~\citep{resmlp} also begin by dividing the input image into patches. It then mixes global information along all patches' spatial and channel dimensions. In a nutshell, different inductive biases and module functions determine different distributions of generated features.

\subsection{Adaptive Knowledge Fusion\label{Three-Level Distillation Paradigm}}
As shown in \cref{tab:Comparisions}, most methods usually apply a two-level paradigm in SAKD~\citep{KD,FitNet,Function}, \ie, T.-S. scheme, to transfer directly the knowledge of teachers to students. Besides, some works apply progressive training strategy~\citep{progressive,multi-teacher} to transfer multi-teacher knowledge in SAKD. However, existing works~\cite{KD,OFA,liu2022cross,wu2024aligning} fall into distillation with a common nature and ignore the unique requirements for specific model pairs, particularly for heterogeneous distillation with significant gaps. A natural question arises: \emph{Can we design a common principle to satisfy varied requirements according to different model pairs?}

Motivated by module connections in FCFD~\cite{Function}, this paper introduces a fusion strategy (called fuse before transfer, FBT), which obeys a common fusion principle but generates different fused models for different T.-S. pairs. Specifically, the knowledge is first fused by merging directly convolution and attention modules derived from both student and teacher module functions and then transferred by training a teacher-fusion-student scheme as follows:

\begin{equation}
\label{eq:FBT-all}
\mathcal{L_\mathrm{{FBT}}} = \mathcal{L}(\mathrm{K}_{\mathrm{t}},\mathrm{K}_{\mathrm{s}}) + \mathcal{L}(\mathrm{K}_{\mathrm{t}},\mathrm{K}_{\mathrm{f}}) + \mathcal{L}(\mathrm{K}_{\mathrm{f}},\mathrm{K}_{\mathrm{s}}),
\end{equation}
where $\mathcal{L}$ is our loss (details in \cref{eq:FBT-Loss}). $\mathrm{K}_{\mathrm{t}}$, $\mathrm{K}_{\mathrm{f}}$, and $\mathrm{K}_{\mathrm{s}}$ denote the knowledge of the T., F., and S. model respectively.

% As a bridge, our fusion model successfully mitigates the gaps between cross-architecture inductive biases and module functions. For instance, in \cref{tab:hybridperformance} and \cref{fig:finalhrbrid}, our fusion model has a middle performance and feature appearances between the students and teachers.

\noindent \textbf{Fusion Strategy.} Our fusion connects the CNN modules and MSA/MLP modules derived from the students and teachers with a local-to-global (L2G) feature projector as shown in \cref{fig:overall}. Formulary, the logits output of our fusion can be described as follows:
\begin{equation}
\label{eq:hybridforward}
p_{\mathrm{f}}(x) = \emph{fc}_{\mathrm{m}} ~\circ \mathrm{S}_{\mathrm{m}}^{4} ~\circ
(\mathrm{MSA} ~\circ
\mathrm{PE}) ~\circ
\mathrm{S}_{\mathrm{c}}^{3} ~\circ \mathrm{S}_{\mathrm{c}}^{2} ~\circ \mathrm{S}_{\mathrm{c}}^{1}(x),
\end{equation}
where $x$ is the input image, $\mathrm{S}_{\mathrm{c}}^{i}$ denotes CNN models, $\mathrm{S}_{\mathrm{m}}^{i}$ denotes MSAs/MLPs, and $fc_{\mathrm{m}}$ denotes the fully-connected layers of MSAs/MLPs. To connect CNN and MSA/MLP modules, we propose an L2G module that includes a patch embedding~\citep{vit} to convert the features into the required dimensions of subsequent MSA/MLP modules. Besides, to capture the long-distance dependency, our L2G also includes an MSA block to project the local features from CNN models to global receptive fields. For simplicity, the MSA module is a Swin block~\citep{swin} in this paper.  \textbf{Note that the fusion is also CNN-MSA/MLP when the teachers are CNNs and student-teacher when the model pairs are homogeneous} (details in our Appendix).

The following considerations drive the design of our fusion: (1) Inductive biases of CNNs and MSAs/MLPs are complementary and hybrid CNN-MSA/MLP models demonstrate good performance~\citep{dovitlikecnn,howvitworks,li2023convmlp}. Therefore, we replace the CNN blocks at the end of a baseline CNN model with MSA/MLP blocks which can benefit from both the local features in the early stages and the global information exchanges in the last stages. For example, the final feature appearances of the fusion are converted from the local CNN features to global receptive fields in \cref{fig:finalhrbrid}. 
(2) Different models have different module functions, which can be aligned implictly by connecting different module functions in a single pipeline~\citep{Function}. Therefore, we form our fusion by using CNN/MSA/MLP modules derived from students and teachers. As shown in \cref{fig:overall}, our fusion is mainly composed of weight-sharing modules $\mathrm{S}_{\mathrm{c}}^{1\rightarrow3}$ and $\mathrm{S}_{\mathrm{m}}^{4}$, which not only unifies different module functions but also introduces negligible additional learnable parameters. Besides, weight-sharing modules also 
ensure the bridge role of our fusion model in \cref{tab:hybridperformance}. 
(3) While alternately using CNN and MSA modules can improve the performance of the fusion~\citep{howvitworks}, we only add one stage of MSA/MLP models following the first three stages of CNN models as illustrated in \cref{eq:hybridforward}, which keeps both simplicity and adaption for different model pairs (details discussed in \cref{discussions}).

\begin{table*}[!t]
\resizebox{\textwidth}{!}{
\begin{tabular}{ccccccccccccc}
\hline \hline
\multirow{2}{*}{Teacher} & \multicolumn{1}{c|}{\multirow{2}{*}{Student}} & \multicolumn{2}{c|}{From Scratch}      & \multicolumn{4}{c|}{feature-based~\cite{FitNet,CC,RKD,CRD}}                     & \multicolumn{3}{c|}{logits-based~\cite{KD,DKD,DIST}}          & \multicolumn{2}{c}{CAKD} \\ \cline{3-13} 
                         & \multicolumn{1}{c|}{}                         & Teacher & \multicolumn{1}{c|}{Student} & FitNet & CC    & RKD   & \multicolumn{1}{c|}{CRD}   & KD    & DKD   & \multicolumn{1}{c|}{DIST}  & OFA   & \cellcolor[HTML]{C0C0C0}\textbf{our FBT} \\ \hline
\multicolumn{13}{l}{\textit{CNN-based students}}                                                                                                                                                                                            \\ \hline
Swin-T                   & \multicolumn{1}{c|}{ResNet18}                 & 89.26   & \multicolumn{1}{c|}{74.01}   & 78.87  & 74.19 & 74.11 & \multicolumn{1}{c|}{77.63} & 78.74 & 80.26 & \multicolumn{1}{c|}{77.75} & \underline{80.54} & \cellcolor[HTML]{C0C0C0}\textbf{81.61}        \\
ViT-S                    & \multicolumn{1}{c|}{ResNet18}                 & 92.04   & \multicolumn{1}{c|}{74.01}   & 77.71  & 74.26 & 73.72 & \multicolumn{1}{c|}{76.60} & 77.26 & 78.10 & \multicolumn{1}{c|}{76.49} & \underline{80.15} & \cellcolor[HTML]{C0C0C0}\textbf{81.93}        \\
Mixer-B/16               & \multicolumn{1}{c|}{ResNet18}                 & 87.29   & \multicolumn{1}{c|}{74.01}   & 77.15  & 74.26 & 73.75 & \multicolumn{1}{c|}{76.42} & 77.79 & 78.67 & \multicolumn{1}{c|}{76.36} & \underline{79.39} & \cellcolor[HTML]{C0C0C0}\textbf{81.90}        \\
Swin-T                   & \multicolumn{1}{c|}{MobileNetV2}              & 89.26   & \multicolumn{1}{c|}{73.68}   & 74.28  & 71.19 & 69.00 & \multicolumn{1}{c|}{79.80} & 74.68 & 71.07 & \multicolumn{1}{c|}{72.89} & \underline{80.98} & \cellcolor[HTML]{C0C0C0}\textbf{81.28}        \\
ViT-S                    & \multicolumn{1}{c|}{MobileNetV2}              & 92.04   & \multicolumn{1}{c|}{73.68}   & 73.54  & 70.67 & 68.46 & \multicolumn{1}{c|}{78.14} & 72.77 & 69.80 & \multicolumn{1}{c|}{72.54} & \underline{78.45} & \cellcolor[HTML]{C0C0C0}\textbf{82.10}        \\
Mixer-B/16               & \multicolumn{1}{c|}{MobileNetV2}              & 87.29   & \multicolumn{1}{c|}{73.68}   & 73.78  & 70.73 & 68.95 & \multicolumn{1}{c|}{78.15} & 73.33 & 70.20 & \multicolumn{1}{c|}{73.26} & \underline{78.78} & \cellcolor[HTML]{C0C0C0}\textbf{80.83}        \\ \hline
\multicolumn{13}{l}{\textit{MSA-based students}}                                                                                                                                                                                            \\ \hline
ConvNeXt-T               & \multicolumn{1}{c|}{DeiT-T}                   & 88.41   & \multicolumn{1}{c|}{68.00}      & 60.78  & 68.01 & 69.79 & \multicolumn{1}{c|}{65.94} & 72.99 & 74.60 & \multicolumn{1}{c|}{73.55} & \underline{75.76} & \cellcolor[HTML]{C0C0C0}\textbf{79.57}        \\
Mixer-B/16               & \multicolumn{1}{c|}{DeiT-T}                   & 87.29   & \multicolumn{1}{c|}{68.00}      & 71.05  & 68.13 & 69.89 & \multicolumn{1}{c|}{65.35} & 71.36 & 73.44 & \multicolumn{1}{c|}{71.67} & \underline{73.90}  & \cellcolor[HTML]{C0C0C0}\textbf{74.40}        \\
ConvNeXt-T               & \multicolumn{1}{c|}{Swin-P}                   & 88.41   & \multicolumn{1}{c|}{72.63}   & 24.06  & 72.63 & 71.73 & \multicolumn{1}{c|}{67.09} & 76.44 & 76.8  & \multicolumn{1}{c|}{76.41} & \underline{78.32} & \cellcolor[HTML]{C0C0C0}\textbf{80.73}        \\
Mixer-B/16               & \multicolumn{1}{c|}{Swin-P}                   & 87.29   & \multicolumn{1}{c|}{72.63}   & 75.2   & 73.32 & 70.82 & \multicolumn{1}{c|}{67.03} & 75.93 & 76.39 & \multicolumn{1}{c|}{75.85} & \underline{76.65} & \cellcolor[HTML]{C0C0C0}\textbf{78.44}        \\ \hline
\multicolumn{13}{l}{\textit{MLP-based students}}                                                                                                                                                                                            \\ \hline
ConvNeXt-t               & \multicolumn{1}{c|}{ResMLP-S12}               & 88.41   & \multicolumn{1}{c|}{66.56}   & 45.47  & 67.70 & 65.82 & \multicolumn{1}{c|}{63.35} & 72.25 & 73.22 & \multicolumn{1}{c|}{71.93} & \underline{75.21} & \cellcolor[HTML]{C0C0C0}\textbf{78.03}        \\
Swin-T                   & \multicolumn{1}{c|}{ResMLP-S12}               & 89.26   & \multicolumn{1}{c|}{66.56}   & 63.12  & 68.37 & 64.66 & \multicolumn{1}{c|}{61.72} & 71.89 & 72.82 & \multicolumn{1}{c|}{11.05} & \underline{73.58} & \cellcolor[HTML]{C0C0C0}\textbf{77.20}        \\ \hline
\multicolumn{2}{c|}{Average Improvements}                                &         & \multicolumn{1}{c|}{}        & $-$5.21  & $-$0.33 & $-$1.39 & \multicolumn{1}{c|}{$-$0.02} & $+$3.12 & $+$3.16 & \multicolumn{1}{c|}{$-$2.31} & \underline{$+$6.19} & \cellcolor[HTML]{C0C0C0}\textbf{$+$8.38}     \\ \hline \hline   
\end{tabular}}
\caption{\textbf{Top-1 accuracy (\%) on CIFAR100.}    All baseline results are from the paper or code of OFA~\citep{OFA}. Swin-P is a modified version of Swin-T\citep{swin} from OFA~\citep{OFA}. \textbf{Bold} denotes the best results and the second-best results are underlined.\label{table:cifar100}}
\vspace{-0.2cm}
\end{table*}

\subsection{Spatial-Agnostic Knowledge Supervision.\label{lossfunction}} As shown in \cref{fig:overall}, we only transfer the final features after average pooling and the logits for the following reasons. (1) Due to the weight-sharing between our fused model and T.-S., it combines actually different inductive biases only in the final features, not early and middle features. (2) As shown in \cref{fig:finalhrbrid} and \cref{fig:teaser}, the final features of different models are also very different in spatial, so we smooth them by average pooling to mitigate the spatial gaps. The knowledge in \cref{eq:FBT-all} is formulated by $\mathrm{K}_{i} = \{f_{i},p_{i}\}$, $i = \mathrm{t}, \mathrm{f}, \mathrm{s}$, where $f_{i}$ and $p_{i}$ denote the final features embeddings after average pooling and the output logits.

In this paper, we use spatial-agnostic InfoNCE loss $\mathcal{L}_{\mathrm{InfoNCE}}$~\citep{he2020momentum,CRD} and OFA loss $\mathcal{L}_{\mathrm{OFA}}$ to supervise the transfer of features and logits respectively, motivated by the following observations. (1) Wildly used MSE loss computes the pixel-wise metrics that are suitable for features having similar spatial information, but it will fail when the features are very spatially different (\eg, FitNet with MSE loss gets only 24.06\% top-1 accuracy when the teacher is ConvNeXt-T and the student is Swin-P in CIFAR100 \cref{table:cifar100}). Consequently, we use a spatial-agnostic loss InfoNCE~\citep{he2020momentum,CRD} to transfer the structural information of feature embeddings~\citep{CRD}, which captures complex interdependencies of features without spatial information. (2) As demonstrated in ~\citep{OFA}, the different inductive bias leads models to variant logit spaces. 
% For example, local CNN models are more suitable for small objects, but global MSA/MLP models are more suitable for large objects. 
Therefore, $\mathcal{L}_{\mathrm{OFA}}$ enhances the information of the target class by adding a modulating parameter $\gamma$ to the original KD loss, which prevents the student from learning incorrect information of the teacher. In a nutshell, our loss $\mathcal{L}$ is suitable for any representation distillation (\eg, the superior consistently performance in \cref{table:cifar100}, \cref{table:imagenet} and \cref{tab:SAKD}):  

\begin{equation}
\label{eq:FBT-Loss}
\left\{
\begin{aligned}
&\mathcal{L}_{\mathrm{OFA}}(p_{\mathrm{t}},p_{\mathrm{s}}) = (1 + p_{\mathrm{t}}^{\hat{c}})^{\gamma} ~\mathrm{log} (\frac{p_{\mathrm{t}}^{\hat{c}}}{p_{\mathrm{s}}^{\hat{c}}}) + \sum_{i=1, i \neq \hat{c}}^{C} p_{\mathrm{t}}^{c}~ \mathrm{log} (\frac{p_{\mathrm{t}}^{c}}{p_{\mathrm{s}}^{c}}),
\\  
&\mathcal{L}_{\mathrm{InfoNCE}}(f_{\mathrm{s}},f_{\mathrm{t}}) = \mathrm{-log}{\frac{\mathrm{exp}(f_{\mathrm{s}} \cdot f_{\mathrm{t}}^{+} / \tau_{2})}{\sum_{i=0}^{F_{\mathrm{t}}}\mathrm{exp}(f_{\mathrm{s}} \cdot f_{\mathrm{t}}^{i}/\tau_{2})}},
\end{aligned}
\right.
\end{equation}

For each T.-S. pair, we transfer knowledge by $\mathcal{L}(\mathrm{K}_{\mathrm{t}},\mathrm{K}_\mathrm{s}) = \mathcal{L}_{\mathrm{InfoNCE}}(f_{\mathrm{t}},f_{\mathrm{s}}) + \mathcal{L}_{\mathrm{OFA}}(p_{\mathrm{t}},p_{\mathrm{s}})$. Firstly, for $\mathcal{L}_{\mathrm{OFA}}$, the $\hat{c}$ and $c$ denote the target class and predicted class of the input image. $C$ is the all classes in the dataset. $\mathcal{L}_{\mathrm{OFA}}$ add a modulating parameter $\gamma$ to enhance the target information when the teacher is not confident about the prediction. 
% When $\gamma = 1$, $\mathcal{L}_{\mathrm{OFA}}$ is equal to $\mathcal{L}_{\mathrm{KD}}$ with temperate $\tau = 1$. 
Secondly, for $\mathcal{L}_{\mathrm{InfoNCE}}$, $f_{\mathrm{s}}$ denotes an encoded student features by average pooling, and $F_{\mathrm{t}}$ is a set of encoded teacher features in a mini-batch. In $F_{\mathrm{t}}$, only one positive sample $f_{\mathrm{t}} ^ {+}$ matches to $f_{\mathrm{s}}$, \ie, the student's and teacher's feature from the same image is a positive pair. The InfoNCE loss is low when the features of student $f_{\mathrm{s}}$ and teacher $f_{\mathrm{t}}^{+}$ are from the same image and high otherwise. This loss has been widely demonstrated for aligning different feature representations~\citep{he2020momentum,CRD}.The temperature parameter $\tau_{2}$ is learnable~\citep{radford2021learning}. Lastly, the entire loss is $\mathcal{L}_{\mathrm{FBT}} = \mathcal{L}(\mathrm{K}_{\mathrm{t}},\mathrm{K}_{\mathrm{s}}) + \mathcal{L}(\mathrm{K}_{\mathrm{t}},\mathrm{K}_{\mathrm{f}}) + \mathcal{L}(\mathrm{K}_{\mathrm{f}},\mathrm{K}_{\mathrm{s}})$, where the formulas of $\mathcal{L}(\mathrm{K}_{\mathrm{t}},\mathrm{K}_{\mathrm{f}})$ and $\mathcal{L}(\mathrm{K}_{\mathrm{f}},\mathrm{K}_{\mathrm{s}})$ is like $\mathcal{L}(\mathrm{K}_{\mathrm{t}},\mathrm{K}_{\mathrm{s}})$.

\section{Experiments}
\subsection{Implementary Details\label{implemendetails}}

\noindent \textbf{Models.}
For a fair comparison, we evaluate our FBT using the same teacher-student pairs employed in OFA\citep{OFA}, including homogeneous distillation and heterogeneous combinations of CNNs, MSAs, and MLPs. 
% Given that the training pipeline for VisionMamba~\citep{mamba} is currently not integrated into the timm library~\citep{rw2019timm}, we don't consider it following OFA~\citep{OFA}. 
Specifically, CNN models include ResNet~\citep{ResNet}, MobileNetv2~\citep{MobileNet}, and ConvNeXt~\citep{liu2022convnet}. MSA models cover ViT, DeiT~\citep{vit,deit}, and Swin~\citep{swin}, while MLP models consist of MLP-Mixer~\citep{mixer} and ResMLP~\citep{resmlp}.

\begin{table*}
\centering

\resizebox{\textwidth}{!}{
\begin{tabular}{ccccccccccccc}
\hline \hline
\multirow{2}{*}{Teacher} & \multicolumn{1}{c|}{\multirow{2}{*}{Student}} & \multicolumn{2}{c|}{From Scratch}      & \multicolumn{4}{c|}{feature-based~\cite{FitNet,CC,RKD,CRD}}                     & \multicolumn{3}{c|}{logits-based~\cite{KD,DKD,DIST}}          & \multicolumn{2}{c}{CAKD} \\ \cline{3-13} 
                         & \multicolumn{1}{c|}{}                         & Teacher & \multicolumn{1}{c|}{Student} & FitNet & CC    & RKD   & \multicolumn{1}{c|}{CRD}   & KD    & DKD   & \multicolumn{1}{c|}{DIST}  & OFA   & \cellcolor[HTML]{C0C0C0}\textbf{our FBT}
  \\ \hline
\multicolumn{13}{l}{\textit{CNN-based models}}                                                                                                                                                                                                                               \\ \hline
DeiT-T                    & \multicolumn{1}{c|}{ResNet18}                  & 72.17 & \multicolumn{1}{c|}{69.75} & 70.44  & 69.77 & 69.47 & \multicolumn{1}{c|}{69.25} & 70.22 & 69.39 & \multicolumn{1}{c|}{70.64} & \multicolumn{1}{c|}{\underline{71.01}} & \cellcolor[HTML]{C0C0C0}\textbf{71.22}  \\
Swin-T                    & \multicolumn{1}{c|}{ResNet18}                  & 81.38 & \multicolumn{1}{c|}{69.75} & 71.18  & 70.07 & 68.89 & \multicolumn{1}{c|}{69.09} & 71.14 & 71.10  & \multicolumn{1}{c|}{70.91} & \multicolumn{1}{c|}{\underline{71.76}} & \cellcolor[HTML]{C0C0C0}\textbf{72.21} \\
Mixer-B/16                & \multicolumn{1}{c|}{ResNet18}                  & 76.62 & \multicolumn{1}{c|}{69.75} & 70.78  & 70.05 & 69.46 & \multicolumn{1}{c|}{68.4}  & 70.89 & 69.89 & \multicolumn{1}{c|}{70.66} & \multicolumn{1}{c|}{\underline{71.38}} & \cellcolor[HTML]{C0C0C0}\textbf{71.44} \\
DeiT-T                    & \multicolumn{1}{c|}{MobileNetV2}               & 72.17 & \multicolumn{1}{c|}{68.87} & 70.95  & 70.69 & 69.72 & \multicolumn{1}{c|}{69.6}  & 70.87 & 70.14 & \multicolumn{1}{c|}{71.08} & \multicolumn{1}{c|}{\underline{71.39}} & \cellcolor[HTML]{C0C0C0}\textbf{71.78} \\
Swin-T                    & \multicolumn{1}{c|}{MobileNetV2}               & 81.38 & \multicolumn{1}{c|}{68.87} & 71.75  & 70.69 & 67.52 & \multicolumn{1}{c|}{69.58} & 72.05 & 71.71 & \multicolumn{1}{c|}{71.76} & \multicolumn{1}{c|}{\underline{72.32}} & \cellcolor[HTML]{C0C0C0}\textbf{72.54} \\
Mixer-B/16                & \multicolumn{1}{c|}{MobileNetV2}               & 76.62 & \multicolumn{1}{c|}{68.87} & 71.59  & 70.79 & 69.86 & \multicolumn{1}{c|}{68.89} & 71.92 & 70.93 & \multicolumn{1}{c|}{71.74} & \multicolumn{1}{c|}{\underline{72.12}} & \cellcolor[HTML]{C0C0C0}\textbf{72.31}  \\ \hline
\multicolumn{13}{l}{\textit{MSA-based Models}}                                                                                                                                                                                                                               \\ \hline
ResNet50                  & \multicolumn{1}{c|}{DeiT-T}                    & 80.38 & \multicolumn{1}{c|}{72.17} & \textbf{75.84}  & 72.56 & 72.06 & \multicolumn{1}{c|}{68.53} & 75.10  & 75.6  & \multicolumn{1}{c|}{75.13} & \multicolumn{1}{c|}{\underline{75.73}} & \cellcolor[HTML]{C0C0C0}75.64     \\
ConvNeXt-T                  & \multicolumn{1}{c|}{DeiT-T}                    & 82.05 & \multicolumn{1}{c|}{72.17} & 70.45  & 73.12 & 71.47 & \multicolumn{1}{c|}{69.18} & 74.00  & 73.95  & \multicolumn{1}{c|}{74.07} & \multicolumn{1}{c|}{\underline{74.41}} & \cellcolor[HTML]{C0C0C0}\textbf{75.26} \\

Mixer-B/16                & \multicolumn{1}{c|}{DeiT-T}                    & 76.62 & \multicolumn{1}{c|}{72.17} & 74.38  & 72.82 & 72.24 & \multicolumn{1}{c|}{68.23} & 74.16 & 72.82 & \multicolumn{1}{c|}{74.22} & \multicolumn{1}{c|}{\underline{74.46}} & \cellcolor[HTML]{C0C0C0}\textbf{75.00}      \\ 
ResNet50                  & \multicolumn{1}{c|}{Swin-N}                    & 80.38 & \multicolumn{1}{c|}{75.53} & 76.83  & 76.05 & 75.90 & \multicolumn{1}{c|}{73.90} & 77.58  & 76.24  & \multicolumn{1}{c|}{77.29} & \multicolumn{1}{c|}{\underline{77.76}} & \cellcolor[HTML]{C0C0C0}\textbf{77.79} \\
ConvNeXt-T                  & \multicolumn{1}{c|}{Swin-N}                    & 82.05 & \multicolumn{1}{c|}{75.53} & 74.81  & 75.79 & 75.48 & \multicolumn{1}{c|}{74.15} & 77.15  & 77.00  & \multicolumn{1}{c|}{77.25} & \multicolumn{1}{c|}{\underline{77.5}} & \cellcolor[HTML]{C0C0C0}\textbf{77.73} \\
Mixer-B/16                & \multicolumn{1}{c|}{Swin-N}                    & 76.62 & \multicolumn{1}{c|}{75.53} & 76.17  & 75.81 & 75.52 & \multicolumn{1}{c|}{73.38} & 76.26 & 75.03 & \multicolumn{1}{c|}{76.54} & \multicolumn{1}{c|}{\underline{76.63}} & \cellcolor[HTML]{C0C0C0}\textbf{76.87}      \\
\hline
\multicolumn{13}{l}{\textit{MLP-based models}}                                                                                          \\ \hline
ConvNeXt-T                  & \multicolumn{1}{c|}{ResMLP-S12}                & 82.05 & \multicolumn{1}{c|}{76.65} & 74.69  & 75.79 & 75.28 & \multicolumn{1}{c|}{73.57} & 76.87 & 77.23 & \multicolumn{1}{c|}{77.24} & \multicolumn{1}{c|}{\underline{77.26}} & \cellcolor[HTML]{C0C0C0}\textbf{77.33}      \\
Swin-T                    & \multicolumn{1}{c|}{ResMLP-S12}                & 81.38 & \multicolumn{1}{c|}{76.65} & 76.48  & 76.15 & 75.1  & \multicolumn{1}{c|}{73.4}  & 76.67 & 76.99 & \multicolumn{1}{c|}{77.25} & \multicolumn{1}{c|}{\underline{77.31}} & \cellcolor[HTML]{C0C0C0}\textbf{77.42}    \\ \hline
\multicolumn{2}{c|}{Average Improvements}                                &         & \multicolumn{1}{c|}{}       & $+$1.00  & $+$0.56 & $-$0.30 & \multicolumn{1}{c|}{$-$1.65} & $+$1.61 & $+$1.05 & \multicolumn{1}{c|}{$+$1.65} & \underline{$+$2.05} & \cellcolor[HTML]{C0C0C0}\textbf{$+$2.31}      \\ \hline \hline
\end{tabular}}
\caption{\textbf{Top-1 accuracy (\%) on ImageNet-1K.} All baseline results are from the paper or code of OFA~\citep{OFA}. Swin-N is a modified version of Swin-T\citep{swin} from OFA~\citep{OFA}. \textbf{Bold} denotes the best results, and the second-best results are underlined.\label{table:imagenet}}

\centering
\vspace{0.3cm}
\resizebox{\textwidth}{!}{
\begin{tabular}{c|cc|cccccccc|cc}
\hline \hline    & T. & S.    & AT~\cite{AT}   & OFD~\cite{OFD}   & CRD~\cite{CRD}  & Review~\cite{reviewkd} & DKD\cite{DKD}   & DIST~\cite{DIST}  & FCFD~\cite{Function}  & OFA~\cite{OFA}   & Ours  \\ \hline
(a)     & 73.31   & 70.66 & 70.69 & 70.81 & 71.17 & 71.61  & 71.70 & 72.07 & \underline{72.24} & 72.10 & \textbf{72.29} \\
(b)    & 76.61   & 68.58 & 69.56 & 71.25 & 71.37 & 72.56  & 72.05 & 73.24 & \underline{73.37} & 73.28 & \textbf{73.45}     \\ \hline \hline
\end{tabular}
}
\caption{\textbf{Results in SAKD on ImageNet-1K.} The teacher and student are ResNet34 and ResNet18 in (a) and are ResNet50 and MobileNet in (b). As shown, our FBT method is still competitive in homogeneous distillation.\label{tab:SAKD}}
\end{table*}

\noindent \textbf{Datasets.}
We use the CIFAR100~\citep{cifar100} and ImageNet-1K dataset~\citep{imagenet} for evaluation.
CIFAR100 consists of 50K training samples and 10K testing samples in a resolution of 32$\times$32, while the ImageNet-1K dataset contains 1.2 million training samples and 50K validation samples with a resolution of 224$\times$224. Since MSAs and MLPs accept image patches as input, we upsample the images in CIFAR100 to the resolution of 224$\times$224~\citep{OFA}.

\noindent \textbf{Baselines.} In line with OFA~\citep{OFA}, we choose several powerful KD methods as our baselines for comparison. Specifically, the feature-based methods include FitNet~\citep{FitNet}, CC~\citep{CC}, RKD~\citep{RKD}, and CRD~\citep{CRD}, while the logits-based methods comprise KD~\citep{KD}, DKD~\citep{DKD}, and DIST~\citep{DIST}. Originally, these methods were designed for SAKD, and thus OFA made some modifications to effectively apply them to CAKD scenarios. Although we are also relevant to ~\cite{liu2022cross,wu2024aligning}, they are not open-sourced and are hard to reproduce in the same experiments.

\noindent \textbf{Training Protocols.} Following the OFA~\cite{OFA}, we utilize SGD optimizer for CNN-based students and AdamW optimizer for MSA- and MLP-based students. All models are trained for 300 epochs on the CIFAR100 dataset. As for the ImageNet-1K dataset, CNNs and MSA/MLP models are trained for 100 epochs and 300 epochs respectively. More details about training schedules are in Appendix A.

\subsection{Main Results\label{mainresults}}

Given extensive cross-architecture teacher-student model pairs,  our FBT consistently achieves the best or most competitive performance on the CIFAR100 (+8.38\% on average Top-1 accuracy) dataset and the ImageNet-1K dataset (+2.31\% on average Top-1 accuracy).    

\begin{table*}[!t]
\centering
\vspace{0.2cm}
\resizebox{0.9\textwidth}{!}{
\begin{tabular}{ccccccccc}
\hline \hline
\multicolumn{1}{c|}{\multirow{3}{*}{Methods}} & \multicolumn{4}{c|}{CIFAR100~\cite{cifar100}}                                                     & \multicolumn{4}{c}{ImageNet~\cite{imagenet}}                                       \\ \cline{2-9} 
\multicolumn{1}{c|}{}                         & T.     & \multicolumn{1}{c|}{S.}       & T.         & \multicolumn{1}{c|}{S.}     & T.     & \multicolumn{1}{c|}{S.}       & T.           & S.         \\ \cline{2-9} 
\multicolumn{1}{c|}{}                         & Swin-T & \multicolumn{1}{c|}{ResNet18} & ConvNeXt-T & \multicolumn{1}{c|}{Swin-P} & Swin-T & \multicolumn{1}{c|}{ResNet18} & ResNet50     & DeiT-T     \\ \hline
\multicolumn{1}{c|}{KD (Baseline)}                      & \multicolumn{2}{c|}{78.74(\textcolor{red}{-2.87})}              & \multicolumn{2}{c|}{76.44(\textcolor{red}{-4.29})}                & \multicolumn{2}{c|}{71.14(\textcolor{red}{-1.04})}              & \multicolumn{2}{c}{75.10(\textcolor{red}{-0.54})} \\
\hline
\multicolumn{9}{l}{The architecture of fused model}                                                                                                                                                   \\ \hline
\multicolumn{1}{c|}{(A) w/o MSA and $\mathrm{S}_{\mathrm{m}}^{4}$ in \cref{eq:hybridforward}}                      & \multicolumn{2}{c|}{75.95(\textcolor{red}{-5.66})}              & \multicolumn{2}{c|}{77.65(\textcolor{red}{-3.18})}                & \multicolumn{2}{c|}{70.86(\textcolor{red}{-1.35})}              & \multicolumn{2}{c}{74.67(\textcolor{red}{-0.97})} \\
\multicolumn{1}{c|}{(B) w/o $\mathrm{S}_{\mathrm{m}}^{4}$ in \cref{eq:hybridforward}}                      & \multicolumn{2}{c|}{77.21(\textcolor{red}{-4.40})}              & \multicolumn{2}{c|}{77.84(\textcolor{red}{-2.89})}                & \multicolumn{2}{c|}{71.78(\textcolor{red}{-0.43})}              & \multicolumn{2}{c}{75.14(\textcolor{red}{-0.50})} \\ \hline
\multicolumn{9}{l}{Loss functions}                                                                 \\ \hline
\multicolumn{1}{c|}{(C) w/o $\mathcal{L}({\mathrm{K}_{\mathrm{t}},\mathrm{K}_{\mathrm{f}}}$) in \cref{eq:FBT-all}}                      & \multicolumn{2}{c|}{25.57(\textcolor{red}{-56.04})}              & \multicolumn{2}{c|}{50.46(\textcolor{red}{-30.27})}                & \multicolumn{2}{c|}{71.34(\textcolor{red}{-0.87})}              & \multicolumn{2}{c}{74.56(\textcolor{red}{-1.08})} \\
\multicolumn{1}{c|}{(D) w/o $\mathcal{L}({\mathrm{K}_{\mathrm{f}},\mathrm{K}_{\mathrm{s}}}$) in \cref{eq:FBT-all}}                      & \multicolumn{2}{c|}{79.01(\textcolor{red}{-2.60})}              & \multicolumn{2}{c|}{79.82(\textcolor{red}{-0.91})}                & \multicolumn{2}{c|}{71.46(\textcolor{red}{-0.75})}              & \multicolumn{2}{c}{74.81(\textcolor{red}{-0.83})} \\
\multicolumn{1}{c|}{(E) w/o $\mathcal{L}({\mathrm{K}_{\mathrm{t}},\mathrm{K}_{\mathrm{s}}}$) in \cref{eq:FBT-all}}                      & \multicolumn{2}{c|}{79.26(\textcolor{red}{-2.35})}              & \multicolumn{2}{c|}{80.17(\textcolor{red}{-0.56})}                & \multicolumn{2}{c|}{71.45(\textcolor{red}{-0.76})}              & \multicolumn{2}{c}{73.92(\textcolor{red}{-1.72})}      \\
\multicolumn{1}{c|}{(F) w/o $\mathcal{L}_{\mathrm{InfoNCE}}$ in \cref{eq:FBT-Loss}}                      & \multicolumn{2}{c|}{80.95(\textcolor{red}{-0.66})}              & \multicolumn{2}{c|}{78.89(\textcolor{red}{-1.84})}                & \multicolumn{2}{c|}{71.47(\textcolor{red}{-0.74})}              & \multicolumn{2}{c}{75.21(\textcolor{red}{-0.43})}      \\ 
\multicolumn{1}{c|}{(G) w/o $\mathcal{L}_{\mathrm{OFA}}$ in \cref{eq:FBT-Loss}}                      & \multicolumn{2}{c|}{77.91(\textcolor{red}{-3.70})}              & \multicolumn{2}{c|}{80.32(\textcolor{red}{-0.41})}                & \multicolumn{2}{c|}{70.37(\textcolor{red}{-1.84})}              & \multicolumn{2}{c}{72.13(\textcolor{red}{-3.51})}      \\ \hline
\multicolumn{1}{c|}{Ours}                     & \multicolumn{2}{c|}{81.61}              & \multicolumn{2}{c|}{80.73}                & \multicolumn{2}{c|}{72.21}              & \multicolumn{2}{c}{75.64}   \\ \hline \hline  
\end{tabular} 
}
\caption{\textbf{Ablation study.} We evaluate the performance by removing some important components of our FBT and loss functions.\label{table:ablation}}
\vspace{-0.3cm}
\end{table*}

\noindent \textbf{Results on CIFAR100.} To evaluate the performance in enough cross-architecture situations, as shown in \cref{table:cifar100}, we conduct extensive experiments in 12 
combinations of heterogeneous T.-S. models. We have the following important observations in this small-scale dataset.

Firstly, feature-based methods exhibit inferior performance on most occasions, \eg, they have negative performance on average improvements, especially when facing the MSA/MLP student models. The reason is that, as discussed in \cref{Preliminaries}, features of cross-architecture models are distinct because of different inductive bias and module functions, in which a naive feature projector struggles to address this dilemma in the small-scale datasets.

Secondly, FitNet~\citep{FitNet} shows very poor performance when the teacher is ConvNeXt-T and the student is Swin-P, while the other feature-based methods obtain relatively normal performance. We believe that this limitation of FitNet stems from its use of the MSE loss to align features in a pixel-wise manner, while other methods solely transfer knowledge from the final feature embeddings or logits. In other words, applying pixel-wise MSE loss may not be suitable for spatially diverse feature maps of student and teacher models, as illustrated in ~\cref{fig:teaser} and Appendix E. 
% Therefore, it is more suitable to transfer final features after smoothing spatial information, the same as our loss function in \cref{eq:FBT-Loss}.

Lastly, OFA~\citep{OFA} yields significant and consistent improvements under all settings. However, these improvements come at the expense of structural feature information by projecting features to logits space. 
% For instance, as shown in \cref{fig:teaser}(e), original feature dimensions exhibit complex interdependencies that would be damaged after projecting the features to logit spaces where each class is more independent~\citep{CRD}. 
In contrast, our framework bridges the cross-architecture representation gaps via a fused model and spatial-agnostic loss applied to spatial-smoothed features. Leveraging the two designs, our FBT achieves the best results in all T.-S. pairs in CAKD, obtaining an average gain of about 2.06\% compared to the recent SOTA method OFA~\citep{OFA} on CIFAR100.

\noindent \textbf{Results on ImageNet-1K.} We also conduct extensive experiments on 14 
combinations of cross-architecture T.-S. models on the large-scale ImageNet-1K dataset. Here, we observe that feature-based methods perform well when handling MSA/MLP students, for instance, distillation with FitNet~\citep{FitNet} when the teacher model is ResNet50 and the student model is DeiT-T. This is opposite to our observations on CIFAR100. We argue that this discrepancy arises due to the data-hungry nature of MLP/MSA models and additional linear feature projectors~\citep{howvitworks}, which are better suited for training on large-scale datasets. Even so, traditional feature-based methods still have negative impacts in some other situations, \eg, FitNet yields 70.45\% (\textcolor{red}{-1.72\%}) when the teacher is ConvNeXt-T and the student is DeiT-T. In a nutshell, even in training with large-scale data, simple feature projectors are not sufficient to align features of cross-architecture T.-S. pairs. 

In this paper, besides the feature projectors L2G, our fused model also includes modules derived from students and frozen teachers. In other words, our fused model achieves a more important task, \ie, aligning the knowledge of student functions to match the frozen teacher functions. Therefore, our FBT leads to superior and stable performance on extensive combinations of cross-architecture models in the small-scale and large-scale dataset. Besides, compared to leveraging four intermediate features of SOTA~\citep{OFA}, our FBT achieves more competitive performance by only leveraging the final features.

\noindent \textbf{Results in SAKD.} As shown in \cref{tab:SAKD}, we compare the distilled results of AT~\citep{AT}, OFD~\citep{OFD}, CRD~\citep{CRD}, Review~\citep{reviewkd}, DKD~\citep{DKD}, DIST~\citep{DIST}, FCFD~\citep{Function} and OFA~\citep{OFA} on ImageNet-1k dataset. Compared to the recent works in homogeneous distillation (FCFD~\citep{Function}) and heterogeneous distillation (OFA~\citep{OFA}), our FBT has a competitive performance. This is because our fusion strategy is also beneficial for aligning different modules in homogeneous distillation. 

% Although the distilled performance of ResNet18 in SAKD is better than CAKD in ~\cref{table:imagenet}, we find that CAKD still improves the distilled student after applying SAKD (multi-teacher progressive distillation~\citep{progressive}, details in our Appendix D). In other words, given a student, heterogeneous teachers have different knowledge compared to homogeneous teachers, which can further improve students. 
% Therefore, our FBT is generic for any T.-S. pairs (including both homogeneous and heterogeneous pairs). 

\subsection{Ablation Study \label{ablation}}

\noindent \textbf{Knowledge Fusion.} In \cref{table:ablation} (A-B), we remove the module $\mathrm{S}_{\mathrm{m}}^{4}$ in (B) and the MSA module in (A), the performance of different teacher-student pairs drops significantly. This demonstrates the power of fusing the inductive biases by adding MSA modules following the CNN modules and fusing the module functions by adding $\mathrm{S}_{\mathrm{m}}^{4}$. Besides, different MSA blocks have different functions for different T.-S. pairs (details in Appendix F), so we use the Swin block as our MSA block in L2G for simplicity.
\begin{table*}[!t]
\centering
\resizebox{.8\textwidth}{!}{
\begin{tabular}{c|ccc|cc}
\hline \hline
Teacher                                                                           & \multicolumn{3}{c|}{fused models with the same length}                                & \multicolumn{2}{c}{The same CNN modules}           \\ \hline
\multirow{5}{*}{\begin{tabular}[c]{@{}c@{}}(A): ViT-S\\ \\ \textcolor{blue}{(B): Swin-T}\end{tabular}} & \multicolumn{1}{c|}{$\mathrm{S}_\mathrm{c}^{1} \rightarrow{\mathrm{S}_\mathrm{m}^{2 \rightarrow fc}}$}              & \multicolumn{1}{c|}{$\mathrm{S}_\mathrm{c}^{1 \rightarrow 2} \rightarrow{\mathrm{S}_\mathrm{m}^{3 \rightarrow fc}}$}             & $\mathrm{S}_\mathrm{c}^{1 \rightarrow 4} \rightarrow{\mathrm{S}_\mathrm{m}^{fc}}$              & \multicolumn{1}{c|}{$\mathrm{S}_\mathrm{c}^{1 \rightarrow 3} \rightarrow{\mathrm{S}_\mathrm{m}^{2 \rightarrow fc}}$}              &  $\mathrm{S}_\mathrm{c}^{1 \rightarrow 3} \rightarrow{\mathrm{S}_\mathrm{m}^{3 \rightarrow fc}}$             \\ \cline{2-6} 
                                                                                     & \multicolumn{1}{c|}{81.5 / \textcolor{blue}{80.56}}  & \multicolumn{1}{c|}{\textbf{82.3} / \textcolor{blue}{79.28}} & 81.15 / \textcolor{blue}{79.27} & \multicolumn{1}{c|}{81.07 / \textcolor{blue}{80.18}} & 82.14 / \textcolor{blue}{78.56} \\ \cline{2-6} 
                                                                                     & \multicolumn{3}{c|}{The same MSA modules}                                              & \multicolumn{2}{c}{Ours}                           \\ \cline{2-6} 
                                                                                     & \multicolumn{1}{c|}{$\mathrm{S}_\mathrm{c}^{1} \rightarrow{\mathrm{S}_\mathrm{m}^{4 \rightarrow fc}}$}              & \multicolumn{1}{c|}{$\mathrm{S}_\mathrm{c}^{1 \rightarrow 2} \rightarrow{\mathrm{S}_\mathrm{m}^{4 \rightarrow fc}}$}             &     $\mathrm{S}_\mathrm{c}^{1 \rightarrow 4} \rightarrow{\mathrm{S}_\mathrm{m}^{4 \rightarrow fc}}$          & \multicolumn{2}{c}{$\mathrm{S}_\mathrm{c}^{1 \rightarrow 3} \rightarrow{\mathrm{S}_\mathrm{m}^{4 \rightarrow fc}}$}                               \\ \cline{2-6} 
                                                                                     & \multicolumn{1}{c|}{80.84 / \textcolor{blue}{80.23}} & \multicolumn{1}{c|}{81.7 / \textcolor{blue}{80.93}} & 80.11 / \textcolor{blue}{79.98} & \multicolumn{2}{c}{81.93 / \textcolor{blue}{\textbf{81.61}}}       \\ \hline \hline          
\end{tabular}
}
\vspace{-0.2cm}
\caption{\textbf{Different fusions.} The student is ResNet18 in CIFAR100. $\mathrm{S}_\mathrm{c}^{1 \rightarrow 2} \rightarrow{\mathrm{S}_\mathrm{m}^{3 \rightarrow fc}}$ denotes we fuse the first two stages of CNN models and the remain parts start from the third stage of MSA models. The others are similar to this definition.\label{table:discussion}\vspace{-0.2cm}}
\end{table*}

\begin{figure*}[!t]
\begin{minipage}{0.5\textwidth}
\resizebox{\textwidth}{!}{
\begin{tabular}{cc|cc|ccc}
\hline \hline
\multirow{2}{*}{Teacher} & \multirow{2}{*}{Student} & \multicolumn{2}{c|}{From Scratch} & \multicolumn{3}{c}{FBT} \\ \cline{3-7} 
                         &                          & T.              & S.              & T.     & F.    & S.    \\ \hline
Swin-T                   & ResNet18                 & 81.38                & 69.75                & 81.38       & \textbf{76.91}         & 72.21      \\
Mixer-B/16               & MobileNetV2              &76.62                 &68.87                 & 76.62       &\textbf{73.85}           &72.31       \\
Mixer-B/16               & DeiT-T                   &76.62                &72.17                 &76.62        &\textbf{76.30}           &75.00       \\
ConvNeXt-T               & ResMLP-S12               & 82.05                &76.65                &82.05        &\textbf{81.20}           &77.33      \\ \hline \hline
\end{tabular}}
\tabcaption{The performance of our fusion model (F.) is between that of the teacher (T.) and student (S.). \label{tab:hybridperformance}}
\end{minipage}
\hspace{0.2cm}
\begin{minipage}{0.45\textwidth}
\includegraphics[width=\linewidth]{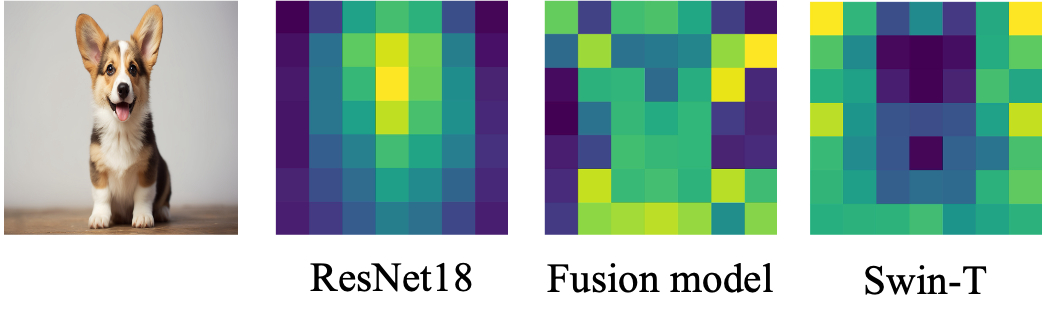}
\vspace{-0.6cm}
\figcaption{The final spatial distribution of the T., F., and S. is different. So we smooth them for alignments in $\mathcal{L}_{\mathrm{InfoNCE}}$. \label{fig:finalhrbrid}}
\end{minipage}
\end{figure*}
\noindent \textbf{Knowledge Transfer.} (1) In \cref{table:ablation} (C), we remove the transfer from the teacher to fused model, \ie, $\mathcal{L}({\mathrm{K}_{\mathrm{t}},\mathrm{K}_{\mathrm{f}}})$, which makes the fused model learn no correct knowledge and then transfers the incorrect knowledge to the students. So the distilled students have poor performance. (2) We remove the transfer from the fused model to students in \cref{table:ablation} 
 (D), \ie, $\mathcal{L}({\mathrm{K}_{\mathrm{f}},\mathrm{K}_{\mathrm{s}}})$. Due to the gaps between heterogeneous students and teachers that are not mitigated without our fusion, the final performance of students is poor too. (3) We remove the transfer from the teacher model to students in \cref{table:ablation} (E), \ie, $\mathcal{L}({\mathrm{K}_{\mathrm{t}},\mathrm{K}_{\mathrm{s}}})$. In this case, although the performance of distilled students is good in some situations, it is not optimal compared to our FBT. This is because our fusion inevitably damages some knowledge from the teachers, and some easy knowledge is more suitable to transfer directly by the T.-S. scheme without a middle bridge. Totally, the proposed fusion strategy and knowledge transfer path is powerful for heterogeneous distillation.

\noindent \textbf{Knowledge Supervision.} In \cref{table:ablation}, we remove the feature loss $\mathcal{L}_{\mathrm{InfoNCE}}$ (F) and the logit loss $\mathcal{L}_{\mathrm{OFA}}$ (G), and different T.-S. pairs have different performance drops on CIFAR100 and ImageNet-1K. For instance, $\mathcal{L}_{\mathrm{OFA}}$ is more important when the teacher is Swin-T and the student is ResNet18 on CIFAR100, while $\mathcal{L}_{\mathrm{InfoNCE}}$ is more important when the teacher is ConvNeXt-T and the student is Swin-P. 
In other words, our $\mathcal{L}_{\mathrm{InfoNCE}}$ and $\mathcal{L}_{\mathrm{OFA}}$ are complementary for different T.-S. pairs, so utilizing them jointly will make various distillations promising. Besides, we demonstrate that MSE loss is not suitable for some T.-S. pairs compared to our INFONCE loss~\citep{CRD} in \cref{mainresults} and Appendix E.

\subsection{Discussion \label{discussions}}

\noindent \textbf{Fuse with different modules.} We compare the performance when we fuse different modules of students and teachers in \cref{table:discussion}. Specifically, we conduct nine different fusions between the student ResNet18 and the teacher (A) ViT-S / (B) Swin-T on CIFAR100. As shown in \cref{table:discussion}, the best result is 82.3\% when the teacher is ViT-S and the fusion is $\mathrm{S}_\mathrm{c}^{1 \rightarrow 2} \rightarrow{\mathrm{S}_\mathrm{m}^{3 \rightarrow fc}}$ and is 81.61\% when the teacher is Swin-T and the fusion is $\mathrm{S}_\mathrm{c}^{1 \rightarrow 3} \rightarrow{\mathrm{S}_\mathrm{m}^{4 \rightarrow fc}}$. For the hybrid fusion, CNN modules and MSA modules are complementary and both play important roles~\citep{howvitworks,dai2021coatnet}. In this paper, although different T.-S pairs have different optimal fusions, we add an MSA/MLP stage following three CNN stages for fusion in most situations, \ie, $\mathrm{S}_\mathrm{c}^{1 \rightarrow 3} \rightarrow{\mathrm{S}_\mathrm{m}^{4 \rightarrow fc}}$,  which ensures simplicity and is adaptive for different model pairs.

\noindent \textbf{The performance and features of our fused model.} Firstly, as shown in \cref{tab:hybridperformance}, our fused model delivers a middle performance between students and teachers, thereby demonstrating its role as a knowledge-fusion bridge. Secondly, as shown in \cref{fig:finalhrbrid}, the final features of our fusion model are global, \ie, our fused model combines the knowledge from different inductive biases and module functions in final feature spaces. Lastly, the features of the student, fusion, and teacher models are spatially different in \cref{fig:finalhrbrid}, so it is reasonable to smooth them before transferring in \cref{eq:FBT-Loss}. Our fusion model plays an important role for bridging the knowledge transfer between T.-S. pairs.

\section{Conclusion}

\noindent \textbf{Limitation and Future Work.} 
(1) It is noteworthy that for certain specific models, such as the extensively studied ResNet18, the performance resulting from distillation by a heterogeneous teacher is inferior to that achieved by a homogeneous teacher. Although our current focus is on the generalizability of various heterogeneous models and yields significant performance improvements in CAKD, a promising avenue for future research may involve additional prior when specific teacher-student pairs are predefined. (2) Our FBT may disrupt heterogeneous features' spatial alignments. This limitation could be mitigated by aligning extra spatial-level distributions (rather than pixel-level). (3) While our approach naturally extends to other tasks, such as object detection and NLP, validating our method across a broader range of domains remains a subject for future investigation.

\noindent \textbf{Conclusion.}
This paper introduces a novel knowledge \textbf{F}usion scheme \textbf{B}efore \textbf{T}ransferring (FBT), which is adaptive according to different model pairs and enhances the efficacy of heterogeneous distillation. Our FBT integrates diverse inductive biases and module functions by fusing CNN/MSA/MLP modules derived from students and teachers, thereby improving the feature transfer among heterogeneous models. Besides, we replace pixel-wise MSE loss with spatial-agnostic loss, which mitigates heterogeneous feature gaps in spatial. Extensive experiments demonstrate that our FBT is more powerful than most homogeneous and heterogeneous methods on CIFAR100 and ImageNet-1K.

\section{Acknowledgement} 
This work is supported by National Natural Science Foundation of China grants under contracts NO.62325111, NO.62306214, and NO.U22B2011.

{
    \small
    \bibliographystyle{ieeenat_fullname}
    \bibliography{main}
}

\end{document}